\documentclass{bmcart}
\usepackage{graphicx}
\usepackage[utf8]{inputenc} 

\startlocaldefs
\endlocaldefs

\begin{document}

\begin{frontmatter}

\begin{fmbox}
\dochead{Technical Report}


\title{Compiling and Processing Historical and Contemporary Portuguese Corpora}


\author[
   addressref={aff1},                   
   corref={aff1},                       
   email={marcos.zampieri@uni-koeln.de}   
]{\inits{MZ}\fnm{Marcos Zampieri} \snm{}}


\address[id=aff1]{
  \orgname{University of Cologne}, 
  \street{Albertus-Magnus Platz},                     %
  \postcode{50923}                                
  \city{Cologne},                              
  \cny{Germany}                                    
}



\end{fmbox}


\begin{abstractbox}

\begin{abstract} 
This technical report describes the framework used for processing three large Portuguese corpora. Two corpora contain texts from newspapers, one published in Brazil and the other published in Portugal. The third corpus is Colonia, a historical Portuguese collection containing texts written between the 16\textsuperscript{th} and the early 20\textsuperscript{th} century. The report presents pre-processing methods, segmentation, and annotation of the corpora as well as indexing and querying methods. Finally, it presents published research papers using the corpora.
\end{abstract}




\end{abstractbox}
%

\end{frontmatter}



\section{Introduction}
\vspace{5mm}

Corpora are great resources for the study of human language. They contain samples of real language use that can be analyzed using text processing software. Corpora provide valuable help for researchers to formulate, confirm or reject hypotheses when studying different aspects of language such as phonetics, syntax or semantics. 

According to several scholars in corpus linguistics  \cite{kennedy1998,mcenerywilson2001,mcenery2011}, one of the main advantages of the use of corpora over native speaker's institution is reproducibility. Findings of corpus-based studies can be replicated much more easily than those obtained relying on human judgments and as a consequence the results obtained are open to objective verification. Leech (1992) \cite{leech1992}, for example, argues that a corpus-based methodology conforms to the scientific method.

The first well-known corpora were compiled for English. Most notable examples include the Brown corpus \cite{francis1979} and the British National Corpus (BNC) \cite{aston1998}. English is still by far the most resource-rich language, but in the last decades several corpora and other language resources have been compiled for other languages as well.

For Portuguese, Linguateca\footnote{http://www.linguateca.pt/} by Santos et al. (2004) \cite{santos2004linguateca} has been the most well-known repository of Portuguese corpora. Through the AC/DC project\footnote{http://www.linguateca.pt/ACDC/} \cite{santos2000} a number of Portuguese corpora of various kinds have been made available to the research community. Colonia, one of the corpora described in this technical report, is also available at Linguateca.

In this report I describe the framework used to compile and annotate three large Portuguese corpora that have been processed at the University of Cologne and released in September 2013. The list of corpora is presented next.

\begin{itemize}
\item {\bf DN-PT}: a journalistic corpus containing texts from the Portuguese newspaper Diário de Notícias published in 2007 and 2008.
\item {\bf FSP-BR}: a journalistic corpus containing texts from the Brazilian newspaper Folha de São Paulo published in 2004.
\item {\bf Colonia}: a diachronic corpus containing full novels published between 16\textsuperscript{th} to the early 20\textsuperscript{th} in Brazil and Portugal \cite{zampieriandbecker13}.
\end{itemize}

\noindent For comparability, the three corpora have been annotated using the same methodology (including tagset) which is described in this report. Copyright restrictions apply to the first two corpora as texts published in these newspapers belong to their respective copyright owners, thus they cannot be freely distributed. Colonia is the only corpus that could be made available to the research community and for this reason it has been used by a number of researchers, as evidenced in this report (see Section 4 for a list of studies using Colonia). 

\vspace{2mm}
\section{Data}
\vspace{5mm}

Compiling and processing corpora is not a trivial task. Best practices include observing several aspects such as balance, sample, and representativeness when collecting data. A corpus is considered to be representative if it takes into account the variability of the population it aims to represent and if the findings obtained based on its content can be generalized to the whole population.

For Colonia\footnote{http://corporavm.uni-koeln.de/colonia/}, the compilation of the material was based on on open repositories such as Domínio Publico\footnote{http://www.dominiopublico.gov.br/} and on existing historical Portuguese corpora, namely the one by Grupo de Morfologia Histórica do Português (GMHP)\footnote{http://www.usp.br/gmhp/Corp.html} from the University of Sao Paulo and Tycho Brahe \cite{galves2010tycho}. A total of 100 texts written and/or published between 1500 and 1948 were collected from these sources and processed using the framework described in this report. 

The DN-PT corpus corpus contains all editions published in 2007 and 2008 by the Lisbon-based newspaper Diário de Notícias were collected whereas FSP-BR contains all texts published by Folha de São Paulo in 2004. 
It should be noted that in the case of these two corpora, even though sampling techniques have been applied, the corpora are not necessarily representative of the whole journalistic genre. For this purpose, samples from multiple newspapers should be collected to account for the variability present in journalistic texts. An example of a journalistic language variety corpus which includes Portuguese and is compiled from multiple newspapers is the DSL Corpus Collection (DSLCC) (Tan et al., 2014) \cite{tan2014merging}.

The total number of tokens in each corpora is presented in Table \ref{tab:all-corpus-size}.

\begin{table}[h]
\centering
\begin{tabular}{c c}
\hline \bf Corpus & \bf Tokens \\ \hline
	DN-PT & 60,845,721 \\
    FSP-BR & 42,258,153\\
    Colonia & 5,157,982 \\
\hline
\end{tabular}
\vspace{2mm}
\caption{\label{tab:all-corpus-size} DN-PT and FSP-BR: Number of Tokens per Corpus.}
\end{table}

\noindent Colonia is publicly available. It can be used through a web-based graphical user interface (GUI) and downloaded in two versions: one containing raw texts and one containing texts with POS annotation. Colonia has been divided into five sub-corpora by century. The number of texts range from 13 texts published in the 16$^{th}$ century and 37 texts published in the 19$^{th}$ century. Table \ref{tab:colonia-corpus-size} presents the number of texts and tokens in each sub-corpus.

\begin{table}[h]
\centering
\begin{tabular}{l c c}
\hline \bf Century & \bf Texts & \bf Tokens \\ \hline
	16$^{th}$ Century & 13 & 399,245 \\
	17$^{th}$ Century & 18 & 709,646 \\
	18$^{th}$ Century & 14 & 425,624 \\
	19$^{th}$ Century & 38 & 2,490,771 \\
	20$^{th}$ Century & 17 & 1,132,696 \\
\hline \bf Total & 100 & \bf 5,157,982 \\
\hline
\end{tabular}
\vspace{2mm}
\caption{\label{tab:colonia-corpus-size} Colonia: Number of Texts and Tokens per Century.}
\end{table}

\vspace{2mm}
\section{Methods}
\vspace{2mm}

The work described in this report uses the tools developed at the Institute for Computational Linguistics of the University of Stuttgart and by Andrew Hardie from the University of Lancaster. For annotation, TreeTagger \cite{schmid94} was used for POS tagging and lemmatization. The Corpus Workbench (CWB) \cite{evert2011twenty} was used for encoding the corpus in a format to be processed by the Corpus Query Processor (CQP) \cite{christ1999ims}. On top of this architecture CQPWeb v3.0.9 \cite{hardie2012cqpweb} was installed in an Apache web server and it is responsible for providing a graphical user interface to CQP queries.

\subsection{Pre-processing, Segmentation, and Annotation}
\vspace{2mm}

The files compiled for the three corpora described in this report, most notably those comprising the two journalistic corpora, contained HTML and XML markup tags. This usually indicates that the data is structured and therefore easier to be parsed but in this case tags were used inconsistently (e.g. opening and not closing, missing attributes, etc.) which made processing these files more challenging. Correcting inconsistent tagging and removing unnecessary tags before segmenting the files (e.g. including sentence and text boundaries) was therefore necessary. For restructuring the files and cleaning tags I tested several HTML and XML parsers available for Python and an XML editor called Stylus Studio.\footnote{http://www.stylusstudio.com/} Some of these parsers did not work at all as they are developed to work on well-formed XML structure. ElementTree\footnote{https://docs.python.org/2/library/xml.etree.elementtree.html}, however, helped me parsing the files to a certain extent. Even so it was necessary to manually check large parts of the files for inconsistencies. This manual step made processing these files very time consuming, but helped to avoid several inconsistencies that could have caused a number of problems in other steps of the processing chain.  

For segmentation, I used Python scripts, regular expressions, and functions available in the Natural Language Toolkit (NLTK) \cite{bird2009natural}. I also performed some high-level normalization on tags and spelling. For Colonia, however, spelling normalization has not been carried out as many texts collected from Tycho Brahe already contained some form of normalized orthography. A post-processing step was carried out on Colonia texts to correct incorrect lemmas after annotation.

\subsection{Annotation}
\vspace{2mm}

For comparability, the three corpora described in this report have been annotated using the same annotation methods. I used TreeTagger \cite{schmid94}, a probabilistic POS tagger, for annotation. Schmid (1994) reports performance of 96.36\% accuracy on annotating the Penn-Treebank. I used the parameter files for Portuguese made available by Pablo Gamallo\footnote{https://gramatica.usc.es/~gamallo/tagger\_intro.htm} and evaluated in Gamallo and Garcia (2013) \cite{gamallo2013freeling}. Gamallo and Garcia (2013) report 96.03\% accuracy on using TreeTagger to annotate a small European Portuguese dataset containing 600 tokens manually annotated for the purpose of evaluation.

The tagset available at Gamallo's Portuguese parameter file contains the following tags:

\vspace{3mm}
\begin{verbatim}
Adjective: ADJ
Adverb: ADV 
Determinant: DET
Cardinal or Ordinal: CARD
Noun: NOM
Pronoun: P
Preposition: PREP
Verb: V
Interjection: I
Punctuation marks: VIRG, SENT
\end{verbatim}
\vspace{3mm}

\noindent The tagset is coarse-grained and morphosyntactic information is not available. Nevertheless, the tagset addresses frequent contractions (prepositions + determiners) and clitics in Portuguese using a few combinations of tags such as:

\vspace{3mm}
\begin{verbatim}
Prepositon + Determiner: PREP+DET
Verb + Prepositon: V+P
\end{verbatim}
\vspace{3mm}

\noindent The aforementioned parameter file by Gamallo was trained on standard contemporary Portuguese. For DN-PT and FSP-BR, I relied solely on the TreeTagger annotation. Lemmatization of the historical texts included in Colonia, however, is more challenging and a few incorrectly attributed lemmas have been correcting in a post-processed step. No systematic correction of POS tags and lemmas has been carried out in any of the three corpora.

Once installed and trained, TreeTagger is fairly easy to use requiring the path to the input file (raw text), the path to the output file (annotated text), the path to the parameter file, and several optional attributes such as $-prob$ to print tag probabilities and $-lemma$ to print lemmas. A sample TreeTagger tagging command is presented next.

\vspace{3mm}

\begin{verbatim}
tree-tagger {options} <parameter file> {<input file> {<output file>}}
\end{verbatim}

\vspace{3mm}

\noindent TreeTagger's output presents one token per line. To be compatible with the input expected by CWB encoding, I used a three-column format which includes the word, POS, and lemma. Segmentation was represented by XML tags and some of them include meta-information as attributes. One example is presented next:

\vspace{3mm}

\begin{verbatim}
<text id="textid">
<s>
Word1	POS1		Lemma1
Word2	POS2		Lemma2
Word3	POS3		Lemma3
</s>
</text>
\end{verbatim}

\vspace{3mm}

\noindent After annotation, the corpus was indexed using the Corpus Workbench (CWB) and then inputted in a CQP installation in a Linux server. This process is described in detail in the next section.

\subsection{Encoding, Indexing, and Querying}
\vspace{2mm}

To be able to perform queries in the corpora, CQP requires the corpus to be encoded in a particular format. The encoding step is carried out by CWB provided that the input text is formatted in the aforementioned one token per line format, also called verticalized text (extension $.vrt$). In the last Section, I showed that TreeTagger produced the correct input required by CWB and no intermediate steps between Treetagger and CWB were necessary. 

To encode the text using CWB it was necessary to follow three steps:

\begin{enumerate}
\item Create a directory to store the binary CWB files (represented by $-d$). 
\item Choose directory to register all encoded corpora (represented by $-R$).
\item Encode the corpus using the $cwb-encode$ choosing positional attributes (represented by $-P$) and structural attributes (represented by $-S$).
\end{enumerate}

\noindent Positional attributes in this case are POS tags and lemmas whereas structural attributes correspond to XML tags which are used to represent, for example, sentence boundaries $<s>$. An example of the $cwb-encode$ command on a file $corpus.vrt$ using POS and lemma as positional attributes and sentence boundaries as structural attribute is presented next.

\vspace{3mm}

\begin{verbatim}
$ cwb-encode -d path corpus.vrt -R path -P pos -P lemma -S s
\end{verbatim}

\vspace{3mm}

\noindent The last steps carried out before the corpora could be used in CQP is indexing and compressing. For that I used the CWB/PERL interface that made indexing and compressing straightforward and quick using $cwb-make$ and the name of the corpus.

Following this procedure the corpora were ready to be used by CQP. The first thing to do to start working with the corpora is to load the corpus on CQP.

\vspace{3mm}

\begin{verbatim}
[no corpus]> COLONIA;
\end{verbatim}

\vspace{3mm}

\noindent After that, queries such as the one presented below can be carried out using the command line.

\vspace{3mm}

\begin{verbatim}
COLONIA> "isso";
\end{verbatim}

\vspace{3mm}

\noindent The previous query will display all occurrences of the word $isso$ in the Colonia corpus. Several options can be changed for better visualization such as the context size. These options are explained in detail in the CQP Tutorial \cite{evert2005cqp} along with many other examples of queries.

After the corpus is loaded, queries can be performed by typing commands through the command line. However, this is not exactly a user-friendly interface. Moreover, this architecture requires users to have access to the server where CQP is installed making it very difficult to release corpora that can be used from outside a particular computer network and/or institution. To improve usability CQPWeb has been developed providing a web-based interface for CQP queries. CQPWeb is designed to be used on top of the architecture presented so far. 

\subsection{Interface}
\vspace{2mm}

Installing the interface was among the most difficult and time consuming procedures described in this report. At the time of the installation, no comprehensive documentation was available for CQPWeb whereas now fortunately a document containing nearly 100 pages is available at the project's website. CQPWeb developers acknowledge that installing the tool is not easy and now they provide a virtual machine with CQPWeb installed called CQPwebInABox\footnote{http://cwb.sourceforge.net/cqpweb.php}. I installed and configured CQPWeb in an Ubuntu Linux virtual machine turned into a web server running Apache, MySQL, and the necessary software required to make CQPWeb work. A dedicated virtual machine has been allocated specially for this corpus compilation project so it was necessary to install all software (including operating systems) from scratch.

\begin{figure}[!ht]
\centering
\includegraphics[width=0.95\textwidth]{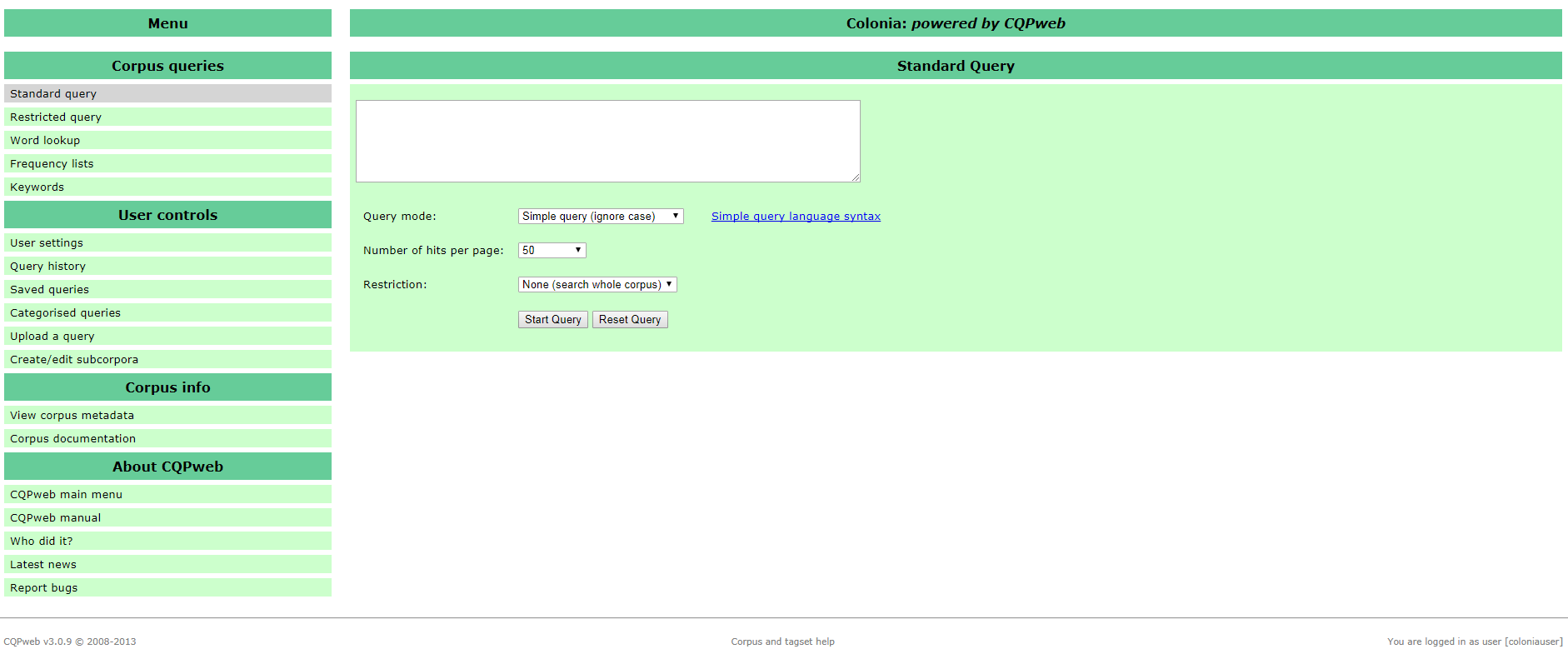}
\caption{A Screenshot of the Standard Query Window of CQPWeb with the Colonia Corpus.}
\label{fig:colonia}
\end{figure}

Once the installation and configuration was complete CQPWeb provides a very user-friendly and easy to use interface for CQP queries. In Figure \ref{fig:colonia} a screenshot of the standard query interface to the Colonia corpus is presented.

\begin{figure}[!ht]
\centering
\includegraphics[width=0.95\textwidth]{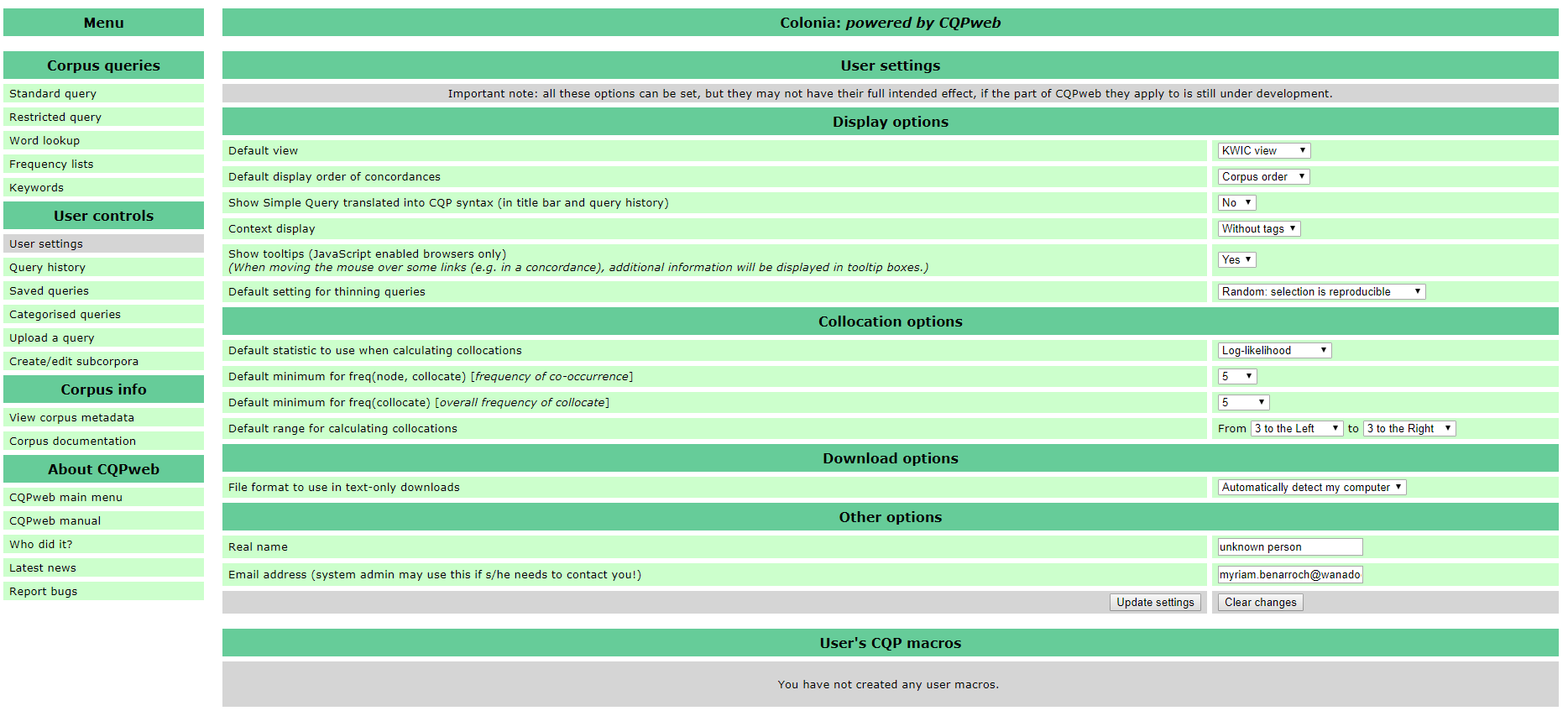}
\caption{A Screenshot of the User Settings in CQPWeb.}
\label{fig:colonia2}
\end{figure}

\noindent On the left hand side the user finds the `Corpus queries' menu with links to calculate frequency lists and keywords. The user is able to change settings and create/edit sub-corpora directly at CQPWeb using the links in the `User controls' menu. By clicking the link `User settings' in the `User controls' menu the user is able to customize several as displayed in Figure \ref{fig:colonia2}.

\section{Applications}
\label{sec:applications}
\vspace{5mm}

To the best of my knowledge,\footnote{This section has been last updated in April 2017.} the corpora presented in this report have been used for a variety of purposes ranging from studies in various topics in linguistics to NLP research. Colonia has been used extensively because it is an open resource that can be used through the CQPWeb interface or downloaded in two versions: raw text or annotated with POS information. After its compilation Colonia has been made available at Corpus Eye\footnote{http://corp.hum.sdu.dk/cqp.pt.html} and Linguateca\footnote{http://www.linguateca.pt/acesso/corpus.php?corpus=COLONIA} with different annotation methods.

A summary of the studies published using the corpora is presented next.

\begin{itemize}
\item {\bf DN-PT and FSP-BR}
\begin{itemize}
\item Language variety identification: There was no standard language variety identification dataset available until the compilation of the DSL Corpus Collection (DSLCC) \cite{tan2014merging}. Samples from these two corpora were used to train computational methods to discriminate between Brazilian and European Portuguese texts  \cite{zampieri2012automatic,zampieri2012classifying,zampieri2013using} as well as to train VarClass \cite{zampieri2014varclass}, an open source language identification tool.
\end{itemize}
\item {\bf Colonia}
\begin{itemize}
\item Semantics: The study by Santos and Mota (2015) \cite{santos2015admiraccao} investigate emotions using the framework by Maia (1994) \cite{maia1994contribution}. The corpora available at the AC/DC project by Linguateca (including Colonia) have been used for this purpose.
\item Dictionary building: Historical dictionaries have been compiled for Portuguese \cite{junior2009building,candido2008procorph} using other corpora. Taking advantage of the annotation carried out using the parser Palavras \cite{bick2000parsing}, the study by Bick and Zampieri (2016) \cite{bick2016grammatical} presents a diachronic dictionary of Portuguese compiled based on the data available in Colonia.
\item Diachronic morphology: A few studies on the diachronic development of Portuguese have used the Colonia corpus \cite{tang2013quantifying,nevins2015rise}. 
\item Temporal text classification: A number of studies on temporal text classification have used Colonia to train systems to predict the publication date of Portuguese texts. Published studies focused on quantifying stylistic changes across centuries using readability metrics \cite{vstajner2013stylistic}, on dating documents using a ranking method \cite{niculae2014temporal}, and on investigating
language change across centuries and in intervals of 50 years \cite{zampieri2016modeling}.
\end{itemize}
\end{itemize}

\vspace{2mm}
\section{Conclusion and Future Work}
\vspace{5mm}

Processing large corpora is time consuming but a rewarding endeavor. Colonia has been used in several research projects, listed in Section \ref{sec:applications}, which compensates the time invested in its compilation. Unfortunately, due to copyright restrictions the two contemporary journalistic corpora, DN-PT and FSP-BR, cannot be used outside of the University of Cologne.

A few directions that the work presented in this report may take in the next few years are:

\begin{itemize}
\item Compiling comparable journalistic corpora for other varieties of Portuguese. These corpora could be annotated and made available using the same framework described in this report resulting a comprehensive account of Portuguese varieties similar to what the Varitext platform offers for French \cite{diwersy2014varitext}. A step in this direction is version 2.1 of the DSLCC which contains texts from Portuguese Newspapers publisher in Macau.\footnote{See \cite{zampieri2016computational} for a computational analysis of the language used in Portuguese newspapers from Macau.}
\item Evaluating the performance of TreeTagger on these corpora using manually annotated gold standard data. This is particularly interesting in the case of Colonia as it is well-known that diachronic language variation, spelling in particular, influences the performance of annotation tools and parsers. Methods described in Hendrickx and Marquilhas (2011) \cite{hendrickx2011old} could be applied to normalize spelling in Colonia.
\item Including fine-grained annotation for verbs and other grammatical categories. The TreeTagger parameter file provided by Pablo Gamallo is trained on a coarse-grained tagset presented in the Section 3 of this report. A version of Colonia available at Corpus Eye, for example, contains morphosyntactic information along with POS tags. A similar annotation could be carried out for the two journalistic corpora as well. 
\end{itemize}

\vspace{2mm}
\section*{Acknowledgements}
\vspace{5mm}

This technical report has been written from February 2012 to September 2013 while I was collecting and processing the corpora. The last (and final) update has been carried out in April 2017.

I would like to thank Sascha Diwersy for the great help provided with CWB and CQP, Charlotte Galves and Mario Viaro for part of the material used to compile Colonia, and Folha de São Paulo for providing us the texts for the FSP-BR corpus.

I further thank Diana Santos and Eckhard Bick for making Colonia available at Linguateca and Corpus Eye respectively.

Finally, I thank all researchers who have been using Colonia for providing valuable feedback and suggestions. 

\vspace{4mm}


\bibliographystyle{bmc-mathphys} 
\bibliography{bmc_article}      


\begin{thebibliography}{36}
\ifx \bisbn   \undefined \def \bisbn  #1{ISBN #1}\fi
\ifx \binits  \undefined \def \binits#1{#1}\fi
\ifx \bauthor  \undefined \def \bauthor#1{#1}\fi
\ifx \batitle  \undefined \def \batitle#1{#1}\fi
\ifx \bjtitle  \undefined \def \bjtitle#1{#1}\fi
\ifx \bvolume  \undefined \def \bvolume#1{\textbf{#1}}\fi
\ifx \byear  \undefined \def \byear#1{#1}\fi
\ifx \bissue  \undefined \def \bissue#1{#1}\fi
\ifx \bfpage  \undefined \def \bfpage#1{#1}\fi
\ifx \blpage  \undefined \def \blpage #1{#1}\fi
\ifx \burl  \undefined \def \burl#1{\textsf{#1}}\fi
\ifx \doiurl  \undefined \def \doiurl#1{\textsf{#1}}\fi
\ifx \betal  \undefined \def \betal{\textit{et al.}}\fi
\ifx \binstitute  \undefined \def \binstitute#1{#1}\fi
\ifx \binstitutionaled  \undefined \def \binstitutionaled#1{#1}\fi
\ifx \bctitle  \undefined \def \bctitle#1{#1}\fi
\ifx \beditor  \undefined \def \beditor#1{#1}\fi
\ifx \bpublisher  \undefined \def \bpublisher#1{#1}\fi
\ifx \bbtitle  \undefined \def \bbtitle#1{#1}\fi
\ifx \bedition  \undefined \def \bedition#1{#1}\fi
\ifx \bseriesno  \undefined \def \bseriesno#1{#1}\fi
\ifx \blocation  \undefined \def \blocation#1{#1}\fi
\ifx \bsertitle  \undefined \def \bsertitle#1{#1}\fi
\ifx \bsnm \undefined \def \bsnm#1{#1}\fi
\ifx \bsuffix \undefined \def \bsuffix#1{#1}\fi
\ifx \bparticle \undefined \def \bparticle#1{#1}\fi
\ifx \barticle \undefined \def \barticle#1{#1}\fi
\ifx \bconfdate \undefined \def \bconfdate #1{#1}\fi
\ifx \botherref \undefined \def \botherref #1{#1}\fi
\ifx \url \undefined \def \url#1{\textsf{#1}}\fi
\ifx \bchapter \undefined \def \bchapter#1{#1}\fi
\ifx \bbook \undefined \def \bbook#1{#1}\fi
\ifx \bcomment \undefined \def \bcomment#1{#1}\fi
\ifx \oauthor \undefined \def \oauthor#1{#1}\fi
\ifx \citeauthoryear \undefined \def \citeauthoryear#1{#1}\fi
\ifx \endbibitem  \undefined \def \endbibitem {}\fi
\ifx \bconflocation  \undefined \def \bconflocation#1{#1}\fi
\ifx \arxivurl  \undefined \def \arxivurl#1{\textsf{#1}}\fi
\csname PreBibitemsHook\endcsname

\bibitem{kennedy1998}
\begin{bbook}
\bauthor{\bsnm{Kennedy}, \binits{G.}}:
\bbtitle{{An Introduction to Corpus Linguistics}}.
\bpublisher{Routledge},
\blocation{London and New York}
(\byear{1998})
\end{bbook}
\endbibitem

\bibitem{mcenerywilson2001}
\begin{bbook}
\bauthor{\bsnm{McEnery}, \binits{A.M.}},
\bauthor{\bsnm{Wilson}, \binits{A.}}:
\bbtitle{{Corpus Linguistics: An Introduction}}.
\bpublisher{Edinburgh University Press},
\blocation{Edinburgh}
(\byear{2001})
\end{bbook}
\endbibitem

\bibitem{mcenery2011}
\begin{bbook}
\bauthor{\bsnm{McEnery}, \binits{T.}},
\bauthor{\bsnm{Hardie}, \binits{A.}}:
\bbtitle{{Corpus Linguistics: Method, Theory and Practice}}.
\bpublisher{Cambridge University Press},
\blocation{Cambridge}
(\byear{2011})
\end{bbook}
\endbibitem

\bibitem{leech1992}
\begin{botherref}
\oauthor{\bsnm{Leech}, \binits{G.}}:
{Corpora and Theories of Linguistic Performance}.
Directions in Corpus Linguistics,
105--122
(1992)
\end{botherref}
\endbibitem

\bibitem{francis1979}
\begin{botherref}
\oauthor{\bsnm{Francis}, \binits{W.N.}},
\oauthor{\bsnm{Kucera}, \binits{H.}}:
{Brown Corpus Manual}.
Brown University
(1979)
\end{botherref}
\endbibitem

\bibitem{aston1998}
\begin{bbook}
\bauthor{\bsnm{Aston}, \binits{G.}},
\bauthor{\bsnm{Burnard}, \binits{L.}}:
\bbtitle{{The BNC Handbook: Exploring the British National Corpus with SARA}}.
\bpublisher{Capstone},
\blocation{Edinburgh}
(\byear{1998})
\end{bbook}
\endbibitem

\bibitem{santos2004linguateca}
\begin{bchapter}
\bauthor{\bsnm{Santos}, \binits{D.}},
\bauthor{\bsnm{Sim{\~o}es}, \binits{A.}},
\bauthor{\bsnm{Frankenberg-Garcia}, \binits{A.}},
\bauthor{\bsnm{Pinto}, \binits{A.}},
\bauthor{\bsnm{Barreiro}, \binits{A.}},
\bauthor{\bsnm{Maia}, \binits{B.}},
\bauthor{\bsnm{Mota}, \binits{C.}},
\bauthor{\bsnm{Oliveira}, \binits{D.}},
\bauthor{\bsnm{Bick}, \binits{E.}},
\bauthor{\bsnm{Ranchhod}, \binits{E.}}, \betal:
\bctitle{{Linguateca: Um Centro de Recursos Distribu{\'\i}do para o
  Processamento Computacional da L{\'\i}ngua Portuguesa}}.
In: \bbtitle{Taller de Herramientas Y Recursos Linguisticos Para el Espanol Y
  el Portugues}
(\byear{2004})
\end{bchapter}
\endbibitem

\bibitem{santos2000}
\begin{bchapter}
\bauthor{\bsnm{Santos}, \binits{D.}},
\bauthor{\bsnm{Bick}, \binits{E.}}:
\bctitle{{Providing Internet Access to Portuguese Corpora: The AC/DC Project}}.
In: \bbtitle{Proceedings of LREC}
(\byear{2000})
\end{bchapter}
\endbibitem

\bibitem{zampieriandbecker13}
\begin{bchapter}
\bauthor{\bsnm{Zampieri}, \binits{M.}},
\bauthor{\bsnm{Becker}, \binits{M.}}:
\bctitle{{Colonia: Corpus of Historical Portuguese}}.
In: \bbtitle{Non-Standard Data Sources in Corpus-Based Research}.
\bsertitle{ZSM Studien},
vol. \bseriesno{5}.
\bpublisher{Shaker},
\blocation{Aachen}
(\byear{2013})
\end{bchapter}
\endbibitem

\bibitem{galves2010tycho}
\begin{botherref}
\oauthor{\bsnm{Galves}, \binits{C.}},
\oauthor{\bsnm{Faria}, \binits{P.}}:
{Tycho Brahe Parsed Corpus of Historical Portuguese}.
URL: http://www.tycho.iel.unicamp.br/corpus/
(2010)
\end{botherref}
\endbibitem

\bibitem{tan2014merging}
\begin{bchapter}
\bauthor{\bsnm{Tan}, \binits{L.}},
\bauthor{\bsnm{Zampieri}, \binits{M.}},
\bauthor{\bsnm{Ljube{\v{s}}ic}, \binits{N.}},
\bauthor{\bsnm{Tiedemann}, \binits{J.}}:
\bctitle{{Merging Comparable Data Sources for the Discrimination of Similar
  Languages: The DSL Corpus Collection}}.
In: \bbtitle{Proceedings of the 7th Workshop on Building and Using Comparable
  Corpora (BUCC)}
(\byear{2014})
\end{bchapter}
\endbibitem

\bibitem{schmid94}
\begin{bchapter}
\bauthor{\bsnm{Schmid}, \binits{H.}}:
\bctitle{{Probabilistic Part-of-Speech Tagging Using Decision Trees}}.
In: \bbtitle{Proceedings of International Conference on New Methods in Language
  Processing},
\bconflocation{Manchester, UK}
(\byear{1994})
\end{bchapter}
\endbibitem

\bibitem{evert2011twenty}
\begin{bchapter}
\bauthor{\bsnm{Evert}, \binits{S.}},
\bauthor{\bsnm{Hardie}, \binits{A.}}:
\bctitle{{Twenty-first Century Corpus Workbench: Updating a Query Architecture
  for the New Millennium}}.
In: \bbtitle{Proceedings of Corpus Linguistics}
(\byear{2011})
\end{bchapter}
\endbibitem

\bibitem{christ1999ims}
\begin{botherref}
\oauthor{\bsnm{Christ}, \binits{O.}},
\oauthor{\bsnm{Schulze}, \binits{B.M.}},
\oauthor{\bsnm{Hofmann}, \binits{A.}},
\oauthor{\bsnm{Koenig}, \binits{E.}}:
{The IMS Corpus Workbench: Corpus Query Processor CQP: User's Manual}.
University of Stuttgart
(1999)
\end{botherref}
\endbibitem

\bibitem{hardie2012cqpweb}
\begin{barticle}
\bauthor{\bsnm{Hardie}, \binits{A.}}:
\batitle{Cqpweb—combining power, flexibility and usability in a corpus
  analysis tool}.
\bjtitle{International journal of corpus linguistics}
\bvolume{17}(\bissue{3}),
\bfpage{380}--\blpage{409}
(\byear{2012})
\end{barticle}
\endbibitem

\bibitem{bird2009natural}
\begin{bbook}
\bauthor{\bsnm{Bird}, \binits{S.}},
\bauthor{\bsnm{Klein}, \binits{E.}},
\bauthor{\bsnm{Loper}, \binits{E.}}:
\bbtitle{{Natural Language Processing with Python: Analyzing Text with the
  Natural Language Toolkit}}.
\bpublisher{O'Reilly},
\blocation{Sebastobol (CA)}
(\byear{2009})
\end{bbook}
\endbibitem

\bibitem{gamallo2013freeling}
\begin{botherref}
\oauthor{\bsnm{Gamallo}, \binits{P.}},
\oauthor{\bsnm{Garcia}, \binits{M.}}:
{FreeLing e TreeTagger: Um Estudo Comparativo no {\^A}mbito do Portugu{\^e}s}.
Technical report,
Universidade de Santiago de Compostela
(2013)
\end{botherref}
\endbibitem

\bibitem{evert2005cqp}
\begin{botherref}
\oauthor{\bsnm{Evert}, \binits{S.}}:
{The CQP Query Language Tutorial}.
IMS, University of Stuttgart
(2005)
\end{botherref}
\endbibitem

\bibitem{zampieri2012automatic}
\begin{bchapter}
\bauthor{\bsnm{Zampieri}, \binits{M.}},
\bauthor{\bsnm{Gebre}, \binits{B.G.}}:
\bctitle{{Automatic Identification of Language Varieties: The Case of
  Portuguese}}.
In: \bbtitle{Proceedings of KONVENS},
pp. \bfpage{233}--\blpage{237}
(\byear{2012})
\end{bchapter}
\endbibitem

\bibitem{zampieri2012classifying}
\begin{bchapter}
\bauthor{\bsnm{Zampieri}, \binits{M.}},
\bauthor{\bsnm{Gebre}, \binits{B.G.}},
\bauthor{\bsnm{Diwersy}, \binits{S.}}:
\bctitle{Classifying pluricentric languages: Extending the monolingual model}.
In: \bbtitle{Proceedings of The Fourth Swedish Language Technology Conference
  (SLTC)},
pp. \bfpage{79}--\blpage{80}
(\byear{2012})
\end{bchapter}
\endbibitem

\bibitem{zampieri2013using}
\begin{bchapter}
\bauthor{\bsnm{Zampieri}, \binits{M.}}:
\bctitle{{Using Bag-of-words to Distinguish Similar languages: How Efficient
  are They?}}
In: \bbtitle{Proceedings of the 14th International Symposium on Computational
  Intelligence and Informatics (CINTI)},
pp. \bfpage{37}--\blpage{41}
(\byear{2013})
\end{bchapter}
\endbibitem

\bibitem{zampieri2014varclass}
\begin{bchapter}
\bauthor{\bsnm{Zampieri}, \binits{M.}},
\bauthor{\bsnm{Gebre}, \binits{B.G.}}:
\bctitle{{VarClass: An Open-source Language Identification Tool for Language
  Varieties}}.
In: \bbtitle{Proceedings of Language Resources and Evaluation (LREC)},
pp. \bfpage{3305}--\blpage{3308}
(\byear{2014})
\end{bchapter}
\endbibitem

\bibitem{santos2015admiraccao}
\begin{botherref}
\oauthor{\bsnm{Santos}, \binits{D.}},
\oauthor{\bsnm{Mota}, \binits{C.}}:
{A Admira{\c{c}}{\~a}o {\`a} Luz dos Corpos}.
Oslo Studies in Language
\textbf{7}(1)
(2015)
\end{botherref}
\endbibitem

\bibitem{maia1994contribution}
\begin{botherref}
\oauthor{\bsnm{Maia}, \binits{B.M.H.S.}}:
{A Contribution to the Study of the language of Emotion in English and
  Portuguese}.
PhD thesis,
University of Porto
(1994)
\end{botherref}
\endbibitem

\bibitem{junior2009building}
\begin{barticle}
\bauthor{\bsnm{Junior}, \binits{A.C.}},
\bauthor{\bsnm{Alu{\'\i}sio}, \binits{S.M.}}:
\batitle{{Building a Corpus-based Historical Portuguese Dictionary: Challenges
  and Opportunities}}.
\bjtitle{TAL}
\bvolume{50}(\bissue{2}),
\bfpage{73}--\blpage{102}
(\byear{2009})
\end{barticle}
\endbibitem

\bibitem{candido2008procorph}
\begin{bchapter}
\bauthor{\bsnm{Candido~Jr}, \binits{A.}},
\bauthor{\bsnm{Alu{\'\i}sio}, \binits{S.}}:
\bctitle{{Procorph: um Sistema de Apoio {\`a} Cria{\c{c}}{\~a}o de
  Dicion{\'a}rios Hist{\'o}ricos}}.
In: \bbtitle{Proceedings of the VI Workshop Tecnologias da Informa{\c{c}}{\~a}o
  e da Linguagem Humana (TIL)},
pp. \bfpage{1}--\blpage{6}
(\byear{2008})
\end{bchapter}
\endbibitem

\bibitem{bick2000parsing}
\begin{bbook}
\bauthor{\bsnm{Bick}, \binits{E.}}:
\bbtitle{{The Parsing System ``Palavras'': Automatic Grammatical Analysis of
  Portuguese in a Constraint Grammar Framework}}.
\bpublisher{Aarhus Universitetsforlag},
\blocation{Arhus, Denmark}
(\byear{2000})
\end{bbook}
\endbibitem

\bibitem{bick2016grammatical}
\begin{bchapter}
\bauthor{\bsnm{Bick}, \binits{E.}},
\bauthor{\bsnm{Zampieri}, \binits{M.}}:
\bctitle{{Grammatical Annotation of Historical Portuguese: Generating a
  Corpus-Based Diachronic Dictionary}}.
In: \bbtitle{Proceedings of Text, Speech, and Dialogue (TSD)},
pp. \bfpage{3}--\blpage{11}
(\byear{2016})
\end{bchapter}
\endbibitem

\bibitem{tang2013quantifying}
\begin{barticle}
\bauthor{\bsnm{Tang}, \binits{K.}},
\bauthor{\bsnm{Nevins}, \binits{A.}}:
\batitle{{Quantifying the Diachronic Productivity of Irregular Verbal Patterns
  in Romance}}.
\bjtitle{UCL Working Papers in Linguistics}
\bvolume{25},
\bfpage{289}--\blpage{308}
(\byear{2013})
\end{barticle}
\endbibitem

\bibitem{nevins2015rise}
\begin{barticle}
\bauthor{\bsnm{Nevins}, \binits{A.}},
\bauthor{\bsnm{Rodrigues}, \binits{C.}},
\bauthor{\bsnm{Tang}, \binits{K.}}:
\batitle{{The Rise and Fall of the L-shaped Morphome: Diachronic and
  Experimental Studies}}.
\bjtitle{Probus}
\bvolume{27}(\bissue{1}),
\bfpage{101}--\blpage{155}
(\byear{2015})
\end{barticle}
\endbibitem

\bibitem{vstajner2013stylistic}
\begin{bchapter}
\bauthor{\bsnm{{\v{S}}tajner}, \binits{S.}},
\bauthor{\bsnm{Zampieri}, \binits{M.}}:
\bctitle{{Stylistic Changes for Temporal Text Classification}}.
In: \bbtitle{Proceeding of Text, Speech and Dialogue (TSD)},
pp. \bfpage{519}--\blpage{526}
(\byear{2013})
\end{bchapter}
\endbibitem

\bibitem{niculae2014temporal}
\begin{bchapter}
\bauthor{\bsnm{Niculae}, \binits{V.}},
\bauthor{\bsnm{Zampieri}, \binits{M.}},
\bauthor{\bsnm{Dinu}, \binits{L.P.}},
\bauthor{\bsnm{Ciobanu}, \binits{A.M.}}:
\bctitle{{Temporal Text Ranking and Automatic Dating of Texts}}.
In: \bbtitle{Proceedings of EACL},
pp. \bfpage{17}--\blpage{21}
(\byear{2014})
\end{bchapter}
\endbibitem

\bibitem{zampieri2016modeling}
\begin{bchapter}
\bauthor{\bsnm{Zampieri}, \binits{M.}},
\bauthor{\bsnm{Malmasi}, \binits{S.}},
\bauthor{\bsnm{Dras}, \binits{M.}}:
\bctitle{{Modeling Language Change in Historical Corpora: The Case of
  Portuguese}}.
In: \bbtitle{Proceedings of Language Resources and Evaluation (LREC)},
pp. \bfpage{4098}--\blpage{4104}
(\byear{2016})
\end{bchapter}
\endbibitem

\bibitem{diwersy2014varitext}
\begin{bchapter}
\bauthor{\bsnm{Diwersy}, \binits{S.}}:
\bctitle{{The Varitext Platform and the Corpus des Vari{\'e}t{\'e}s Nationales
  du Fran{\c{c}}ais (CoVaNa-FR) as Resources for the Study of French from a
  Pluricentric Perspective}}.
In: \bbtitle{Proceedings of the First Workshop on Applying NLP Tools to Similar
  Languages, Varieties and Dialects (VarDial)}
(\byear{2014})
\end{bchapter}
\endbibitem

\bibitem{zampieri2016computational}
\begin{bchapter}
\bauthor{\bsnm{Zampieri}, \binits{M.}},
\bauthor{\bsnm{Malmasi}, \binits{S.}},
\bauthor{\bsnm{Sulea}, \binits{O.-M.}},
\bauthor{\bsnm{Dinu}, \binits{L.P.}}:
\bctitle{{A Computational Approach to the Study of Portuguese Newspapers
  Published in Macau}}.
In: \bbtitle{Proceedings of Workshop on Natural Language Processing Meets
  Journalism (NLPMJ)},
pp. \bfpage{47}--\blpage{51}
(\byear{2016})
\end{bchapter}
\endbibitem

\bibitem{hendrickx2011old}
\begin{barticle}
\bauthor{\bsnm{Hendrickx}, \binits{I.}},
\bauthor{\bsnm{Marquilhas}, \binits{R.}}:
\batitle{{From Old Texts to Modern Spellings: An Experiment in Automatic
  Normalisation}}.
\bjtitle{JLCL}
\bvolume{26}(\bissue{2}),
\bfpage{65}--\blpage{76}
(\byear{2011})
\end{barticle}
\endbibitem

\end{thebibliography}

\newcommand{\BMCxmlcomment}[1]{}

\BMCxmlcomment{

<refgrp>

<bibl id="B1">
  <title><p>{An Introduction to Corpus Linguistics}</p></title>
  <aug>
    <au><snm>Kennedy</snm><fnm>G</fnm></au>
  </aug>
  <publisher>London and New York: Routledge</publisher>
  <pubdate>1998</pubdate>
</bibl>

<bibl id="B2">
  <title><p>{Corpus Linguistics: An Introduction}</p></title>
  <aug>
    <au><snm>McEnery</snm><fnm>AM</fnm></au>
    <au><snm>Wilson</snm><fnm>A</fnm></au>
  </aug>
  <publisher>Edinburgh: Edinburgh University Press</publisher>
  <pubdate>2001</pubdate>
</bibl>

<bibl id="B3">
  <title><p>{Corpus Linguistics: Method, Theory and Practice}</p></title>
  <aug>
    <au><snm>McEnery</snm><fnm>T</fnm></au>
    <au><snm>Hardie</snm><fnm>A</fnm></au>
  </aug>
  <publisher>Cambridge: Cambridge University Press</publisher>
  <pubdate>2011</pubdate>
</bibl>

<bibl id="B4">
  <title><p>{Corpora and Theories of Linguistic Performance}</p></title>
  <aug>
    <au><snm>Leech</snm><fnm>G</fnm></au>
  </aug>
  <source>Directions in Corpus Linguistics</source>
  <pubdate>1992</pubdate>
  <fpage>105</fpage>
  <lpage>-122</lpage>
</bibl>

<bibl id="B5">
  <title><p>{Brown Corpus Manual}</p></title>
  <aug>
    <au><snm>Francis</snm><fnm>WN</fnm></au>
    <au><snm>Kucera</snm><fnm>H</fnm></au>
  </aug>
  <source>Brown University</source>
  <pubdate>1979</pubdate>
</bibl>

<bibl id="B6">
  <title><p>{The BNC Handbook: Exploring the British National Corpus with
  SARA}</p></title>
  <aug>
    <au><snm>Aston</snm><fnm>G</fnm></au>
    <au><snm>Burnard</snm><fnm>L</fnm></au>
  </aug>
  <publisher>Edinburgh: Capstone</publisher>
  <pubdate>1998</pubdate>
</bibl>

<bibl id="B7">
  <title><p>{Linguateca: Um Centro de Recursos Distribu{\'\i}do para o
  Processamento Computacional da L{\'\i}ngua Portuguesa}</p></title>
  <aug>
    <au><snm>Santos</snm><fnm>D</fnm></au>
    <au><snm>Sim{\~o}es</snm><fnm>A</fnm></au>
    <au><snm>Frankenberg Garcia</snm><fnm>A</fnm></au>
    <au><snm>Pinto</snm><fnm>A</fnm></au>
    <au><snm>Barreiro</snm><fnm>A</fnm></au>
    <au><snm>Maia</snm><fnm>B</fnm></au>
    <au><snm>Mota</snm><fnm>C</fnm></au>
    <au><snm>Oliveira</snm><fnm>D</fnm></au>
    <au><snm>Bick</snm><fnm>E</fnm></au>
    <au><snm>Ranchhod</snm><fnm>E</fnm></au>
    <au><cnm>others</cnm></au>
  </aug>
  <source>Taller de Herramientas y Recursos Linguisticos para el Espanol y el
  Portugues</source>
  <pubdate>2004</pubdate>
</bibl>

<bibl id="B8">
  <title><p>{Providing Internet Access to Portuguese Corpora: The AC/DC
  Project}</p></title>
  <aug>
    <au><snm>Santos</snm><fnm>D</fnm></au>
    <au><snm>Bick</snm><fnm>E</fnm></au>
  </aug>
  <source>Proceedings of LREC</source>
  <pubdate>2000</pubdate>
</bibl>

<bibl id="B9">
  <title><p>{Colonia: Corpus of Historical Portuguese}</p></title>
  <aug>
    <au><snm>Zampieri</snm><fnm>M</fnm></au>
    <au><snm>Becker</snm><fnm>M</fnm></au>
  </aug>
  <source>Non-Standard Data Sources in Corpus-Based Research</source>
  <publisher>Aachen: Shaker</publisher>
  <series><title><p>ZSM Studien</p></title></series>
  <pubdate>2013</pubdate>
  <volume>5</volume>
</bibl>

<bibl id="B10">
  <title><p>{Tycho Brahe Parsed Corpus of Historical Portuguese}</p></title>
  <aug>
    <au><snm>Galves</snm><fnm>C</fnm></au>
    <au><snm>Faria</snm><fnm>P</fnm></au>
  </aug>
  <source>URL: http://www.tycho.iel.unicamp.br/corpus/</source>
  <pubdate>2010</pubdate>
</bibl>

<bibl id="B11">
  <title><p>{Merging Comparable Data Sources for the Discrimination of Similar
  Languages: The DSL Corpus Collection}</p></title>
  <aug>
    <au><snm>Tan</snm><fnm>L</fnm></au>
    <au><snm>Zampieri</snm><fnm>M</fnm></au>
    <au><snm>Ljube{\v{s}}ic</snm><fnm>N</fnm></au>
    <au><snm>Tiedemann</snm><fnm>J</fnm></au>
  </aug>
  <source>Proceedings of the 7th Workshop on Building and Using Comparable
  Corpora (BUCC)</source>
  <pubdate>2014</pubdate>
</bibl>

<bibl id="B12">
  <title><p>{Probabilistic Part-of-Speech Tagging Using Decision
  Trees}</p></title>
  <aug>
    <au><snm>Schmid</snm><fnm>H.</fnm></au>
  </aug>
  <source>Proceedings of International Conference on New Methods in Language
  Processing</source>
  <publisher>Manchester, UK</publisher>
  <pubdate>1994</pubdate>
</bibl>

<bibl id="B13">
  <title><p>{Twenty-first Century Corpus Workbench: Updating a Query
  Architecture for the New Millennium}</p></title>
  <aug>
    <au><snm>Evert</snm><fnm>S</fnm></au>
    <au><snm>Hardie</snm><fnm>A</fnm></au>
  </aug>
  <source>Proceedings of Corpus Linguistics</source>
  <pubdate>2011</pubdate>
</bibl>

<bibl id="B14">
  <title><p>{The IMS Corpus Workbench: Corpus Query Processor CQP: User's
  Manual}</p></title>
  <aug>
    <au><snm>Christ</snm><fnm>O</fnm></au>
    <au><snm>Schulze</snm><fnm>BM</fnm></au>
    <au><snm>Hofmann</snm><fnm>A</fnm></au>
    <au><snm>Koenig</snm><fnm>E</fnm></au>
  </aug>
  <source>University of Stuttgart</source>
  <pubdate>1999</pubdate>
</bibl>

<bibl id="B15">
  <title><p>CQPweb—combining power, flexibility and usability in a corpus
  analysis tool</p></title>
  <aug>
    <au><snm>Hardie</snm><fnm>A</fnm></au>
  </aug>
  <source>International journal of corpus linguistics</source>
  <publisher>John Benjamins Publishing Company</publisher>
  <pubdate>2012</pubdate>
  <volume>17</volume>
  <issue>3</issue>
  <fpage>380</fpage>
  <lpage>-409</lpage>
</bibl>

<bibl id="B16">
  <title><p>{Natural Language Processing with Python: Analyzing Text with the
  Natural Language Toolkit}</p></title>
  <aug>
    <au><snm>Bird</snm><fnm>S</fnm></au>
    <au><snm>Klein</snm><fnm>E</fnm></au>
    <au><snm>Loper</snm><fnm>E</fnm></au>
  </aug>
  <publisher>Sebastobol (CA): O'Reilly</publisher>
  <pubdate>2009</pubdate>
</bibl>

<bibl id="B17">
  <title><p>{FreeLing e TreeTagger: Um Estudo Comparativo no {\^A}mbito do
  Portugu{\^e}s}</p></title>
  <aug>
    <au><snm>Gamallo</snm><fnm>P</fnm></au>
    <au><snm>Garcia</snm><fnm>M</fnm></au>
  </aug>
  <pubdate>2013</pubdate>
</bibl>

<bibl id="B18">
  <title><p>{The CQP Query Language Tutorial}</p></title>
  <aug>
    <au><snm>Evert</snm><fnm>S</fnm></au>
  </aug>
  <source>IMS, University of Stuttgart</source>
  <pubdate>2005</pubdate>
</bibl>

<bibl id="B19">
  <title><p>{Automatic Identification of Language Varieties: The Case of
  Portuguese}</p></title>
  <aug>
    <au><snm>Zampieri</snm><fnm>M</fnm></au>
    <au><snm>Gebre</snm><fnm>BG</fnm></au>
  </aug>
  <source>Proceedings of KONVENS</source>
  <pubdate>2012</pubdate>
  <fpage>233</fpage>
  <lpage>-237</lpage>
</bibl>

<bibl id="B20">
  <title><p>Classifying Pluricentric Languages: Extending the Monolingual
  Model</p></title>
  <aug>
    <au><snm>Zampieri</snm><fnm>M</fnm></au>
    <au><snm>Gebre</snm><fnm>BG</fnm></au>
    <au><snm>Diwersy</snm><fnm>S</fnm></au>
  </aug>
  <source>Proceedings of The Fourth Swedish Language Technology Conference
  (SLTC)</source>
  <pubdate>2012</pubdate>
  <fpage>79</fpage>
  <lpage>-80</lpage>
</bibl>

<bibl id="B21">
  <title><p>{Using Bag-of-words to Distinguish Similar languages: How Efficient
  are They?}</p></title>
  <aug>
    <au><snm>Zampieri</snm><fnm>M</fnm></au>
  </aug>
  <source>Proceedings of the 14th International Symposium on Computational
  Intelligence and Informatics (CINTI)</source>
  <pubdate>2013</pubdate>
  <fpage>37</fpage>
  <lpage>-41</lpage>
</bibl>

<bibl id="B22">
  <title><p>{VarClass: An Open-source Language Identification Tool for Language
  Varieties}</p></title>
  <aug>
    <au><snm>Zampieri</snm><fnm>M</fnm></au>
    <au><snm>Gebre</snm><fnm>BG</fnm></au>
  </aug>
  <source>Proceedings of Language Resources and Evaluation (LREC)</source>
  <pubdate>2014</pubdate>
  <fpage>3305</fpage>
  <lpage>-3308</lpage>
</bibl>

<bibl id="B23">
  <title><p>{A Admira{\c{c}}{\~a}o {\`a} Luz dos Corpos}</p></title>
  <aug>
    <au><snm>Santos</snm><fnm>D</fnm></au>
    <au><snm>Mota</snm><fnm>C</fnm></au>
  </aug>
  <source>Oslo Studies in Language</source>
  <pubdate>2015</pubdate>
  <volume>7</volume>
  <issue>1</issue>
</bibl>

<bibl id="B24">
  <title><p>{A Contribution to the Study of the language of Emotion in English
  and Portuguese}</p></title>
  <aug>
    <au><snm>Maia</snm><fnm>BMHS</fnm></au>
  </aug>
  <source>PhD thesis</source>
  <publisher>University of Porto</publisher>
  <pubdate>1994</pubdate>
</bibl>

<bibl id="B25">
  <title><p>{Building a Corpus-based Historical Portuguese Dictionary:
  Challenges and Opportunities}</p></title>
  <aug>
    <au><snm>Junior</snm><fnm>AC</fnm></au>
    <au><snm>Alu{\'\i}sio</snm><fnm>SM</fnm></au>
  </aug>
  <source>TAL</source>
  <pubdate>2009</pubdate>
  <volume>50</volume>
  <issue>2</issue>
  <fpage>73</fpage>
  <lpage>-102</lpage>
</bibl>

<bibl id="B26">
  <title><p>{Procorph: um Sistema de Apoio {\`a} Cria{\c{c}}{\~a}o de
  Dicion{\'a}rios Hist{\'o}ricos}</p></title>
  <aug>
    <au><snm>Candido Jr</snm><fnm>A</fnm></au>
    <au><snm>Alu{\'\i}sio</snm><fnm>SM</fnm></au>
  </aug>
  <source>Proceedings of the VI Workshop Tecnologias da Informa{\c{c}}{\~a}o e
  da Linguagem Humana (TIL)</source>
  <pubdate>2008</pubdate>
  <fpage>1</fpage>
  <lpage>-6</lpage>
</bibl>

<bibl id="B27">
  <title><p>{The Parsing System ``Palavras'': Automatic Grammatical Analysis of
  Portuguese in a Constraint Grammar Framework}</p></title>
  <aug>
    <au><snm>Bick</snm><fnm>E</fnm></au>
  </aug>
  <publisher>Arhus, Denmark: Aarhus Universitetsforlag</publisher>
  <pubdate>2000</pubdate>
</bibl>

<bibl id="B28">
  <title><p>{Grammatical Annotation of Historical Portuguese: Generating a
  Corpus-Based Diachronic Dictionary}</p></title>
  <aug>
    <au><snm>Bick</snm><fnm>E</fnm></au>
    <au><snm>Zampieri</snm><fnm>M</fnm></au>
  </aug>
  <source>Proceedings of Text, Speech, and Dialogue (TSD)</source>
  <pubdate>2016</pubdate>
  <fpage>3</fpage>
  <lpage>-11</lpage>
</bibl>

<bibl id="B29">
  <title><p>{Quantifying the Diachronic Productivity of Irregular Verbal
  Patterns in Romance}</p></title>
  <aug>
    <au><snm>Tang</snm><fnm>K</fnm></au>
    <au><snm>Nevins</snm><fnm>A</fnm></au>
  </aug>
  <source>UCL Working Papers in Linguistics</source>
  <publisher>UCL Working Papers in Linguistics</publisher>
  <pubdate>2013</pubdate>
  <volume>25</volume>
  <fpage>289</fpage>
  <lpage>-308</lpage>
</bibl>

<bibl id="B30">
  <title><p>{The Rise and Fall of the L-shaped Morphome: Diachronic and
  Experimental Studies}</p></title>
  <aug>
    <au><snm>Nevins</snm><fnm>A</fnm></au>
    <au><snm>Rodrigues</snm><fnm>C</fnm></au>
    <au><snm>Tang</snm><fnm>K</fnm></au>
  </aug>
  <source>Probus</source>
  <pubdate>2015</pubdate>
  <volume>27</volume>
  <issue>1</issue>
  <fpage>101</fpage>
  <lpage>-155</lpage>
</bibl>

<bibl id="B31">
  <title><p>{Stylistic Changes for Temporal Text Classification}</p></title>
  <aug>
    <au><snm>{\v{S}}tajner</snm><fnm>S</fnm></au>
    <au><snm>Zampieri</snm><fnm>M</fnm></au>
  </aug>
  <source>Proceeding of Text, Speech and Dialogue (TSD)</source>
  <pubdate>2013</pubdate>
  <fpage>519</fpage>
  <lpage>-526</lpage>
</bibl>

<bibl id="B32">
  <title><p>{Temporal Text Ranking and Automatic Dating of Texts}</p></title>
  <aug>
    <au><snm>Niculae</snm><fnm>V</fnm></au>
    <au><snm>Zampieri</snm><fnm>M</fnm></au>
    <au><snm>Dinu</snm><fnm>LP</fnm></au>
    <au><snm>Ciobanu</snm><fnm>AM</fnm></au>
  </aug>
  <source>Proceedings of EACL</source>
  <pubdate>2014</pubdate>
  <fpage>17</fpage>
  <lpage>-21</lpage>
</bibl>

<bibl id="B33">
  <title><p>{Modeling Language Change in Historical Corpora: The Case of
  Portuguese}</p></title>
  <aug>
    <au><snm>Zampieri</snm><fnm>M</fnm></au>
    <au><snm>Malmasi</snm><fnm>S</fnm></au>
    <au><snm>Dras</snm><fnm>M</fnm></au>
  </aug>
  <source>Proceedings of Language Resources and Evaluation (LREC)</source>
  <pubdate>2016</pubdate>
  <fpage>4098</fpage>
  <lpage>-4104</lpage>
</bibl>

<bibl id="B34">
  <title><p>{The Varitext Platform and the Corpus des Vari{\'e}t{\'e}s
  Nationales du Fran{\c{c}}ais (CoVaNa-FR) as Resources for the Study of French
  from a Pluricentric Perspective}</p></title>
  <aug>
    <au><snm>Diwersy</snm><fnm>S</fnm></au>
  </aug>
  <source>Proceedings of the First Workshop on Applying NLP Tools to Similar
  Languages, Varieties and Dialects (VarDial)</source>
  <pubdate>2014</pubdate>
</bibl>

<bibl id="B35">
  <title><p>{A Computational Approach to the Study of Portuguese Newspapers
  Published in Macau}</p></title>
  <aug>
    <au><snm>Zampieri</snm><fnm>M</fnm></au>
    <au><snm>Malmasi</snm><fnm>S</fnm></au>
    <au><snm>Sulea</snm><fnm>OM</fnm></au>
    <au><snm>Dinu</snm><fnm>LP</fnm></au>
  </aug>
  <source>Proceedings of Workshop on Natural Language Processing Meets
  Journalism (NLPMJ)</source>
  <pubdate>2016</pubdate>
  <fpage>47</fpage>
  <lpage>-51</lpage>
</bibl>

<bibl id="B36">
  <title><p>{From Old Texts to Modern Spellings: An Experiment in Automatic
  Normalisation}</p></title>
  <aug>
    <au><snm>Hendrickx</snm><fnm>I</fnm></au>
    <au><snm>Marquilhas</snm><fnm>R</fnm></au>
  </aug>
  <source>JLCL</source>
  <pubdate>2011</pubdate>
  <volume>26</volume>
  <issue>2</issue>
  <fpage>65</fpage>
  <lpage>-76</lpage>
</bibl>

</refgrp>
} 





\newcommand{\BMCxmlcomment}[1]{}

\BMCxmlcomment{

<refgrp>

<bibl id="B1">
  <title><p>{An Introduction to Corpus Linguistics}</p></title>
  <aug>
    <au><snm>Kennedy</snm><fnm>G</fnm></au>
  </aug>
  <publisher>London and New York: Routledge</publisher>
  <pubdate>1998</pubdate>
</bibl>

<bibl id="B2">
  <title><p>{Corpus Linguistics: An Introduction}</p></title>
  <aug>
    <au><snm>McEnery</snm><fnm>AM</fnm></au>
    <au><snm>Wilson</snm><fnm>A</fnm></au>
  </aug>
  <publisher>Edinburgh: Edinburgh University Press</publisher>
  <pubdate>2001</pubdate>
</bibl>

<bibl id="B3">
  <title><p>{Corpus Linguistics: Method, Theory and Practice}</p></title>
  <aug>
    <au><snm>McEnery</snm><fnm>T</fnm></au>
    <au><snm>Hardie</snm><fnm>A</fnm></au>
  </aug>
  <publisher>Cambridge: Cambridge University Press</publisher>
  <pubdate>2011</pubdate>
</bibl>

<bibl id="B4">
  <title><p>{Corpora and Theories of Linguistic Performance}</p></title>
  <aug>
    <au><snm>Leech</snm><fnm>G</fnm></au>
  </aug>
  <source>Directions in Corpus Linguistics</source>
  <pubdate>1992</pubdate>
  <fpage>105</fpage>
  <lpage>-122</lpage>
</bibl>

<bibl id="B5">
  <title><p>{Brown Corpus Manual}</p></title>
  <aug>
    <au><snm>Francis</snm><fnm>WN</fnm></au>
    <au><snm>Kucera</snm><fnm>H</fnm></au>
  </aug>
  <source>Brown University</source>
  <pubdate>1979</pubdate>
</bibl>

<bibl id="B6">
  <title><p>{The BNC Handbook: Exploring the British National Corpus with
  SARA}</p></title>
  <aug>
    <au><snm>Aston</snm><fnm>G</fnm></au>
    <au><snm>Burnard</snm><fnm>L</fnm></au>
  </aug>
  <publisher>Edinburgh: Capstone</publisher>
  <pubdate>1998</pubdate>
</bibl>

<bibl id="B7">
  <title><p>{Linguateca: Um Centro de Recursos Distribu{\'\i}do para o
  Processamento Computacional da L{\'\i}ngua Portuguesa}</p></title>
  <aug>
    <au><snm>Santos</snm><fnm>D</fnm></au>
    <au><snm>Sim{\~o}es</snm><fnm>A</fnm></au>
    <au><snm>Frankenberg Garcia</snm><fnm>A</fnm></au>
    <au><snm>Pinto</snm><fnm>A</fnm></au>
    <au><snm>Barreiro</snm><fnm>A</fnm></au>
    <au><snm>Maia</snm><fnm>B</fnm></au>
    <au><snm>Mota</snm><fnm>C</fnm></au>
    <au><snm>Oliveira</snm><fnm>D</fnm></au>
    <au><snm>Bick</snm><fnm>E</fnm></au>
    <au><snm>Ranchhod</snm><fnm>E</fnm></au>
    <au><cnm>others</cnm></au>
  </aug>
  <source>Taller de Herramientas y Recursos Linguisticos para el Espanol y el
  Portugues</source>
  <pubdate>2004</pubdate>
</bibl>

<bibl id="B8">
  <title><p>{Providing Internet Access to Portuguese Corpora: The AC/DC
  Project}</p></title>
  <aug>
    <au><snm>Santos</snm><fnm>D</fnm></au>
    <au><snm>Bick</snm><fnm>E</fnm></au>
  </aug>
  <source>Proceedings of LREC</source>
  <pubdate>2000</pubdate>
</bibl>

<bibl id="B9">
  <title><p>{Colonia: Corpus of Historical Portuguese}</p></title>
  <aug>
    <au><snm>Zampieri</snm><fnm>M</fnm></au>
    <au><snm>Becker</snm><fnm>M</fnm></au>
  </aug>
  <source>Non-Standard Data Sources in Corpus-Based Research</source>
  <publisher>Aachen: Shaker</publisher>
  <series><title><p>ZSM Studien</p></title></series>
  <pubdate>2013</pubdate>
  <volume>5</volume>
</bibl>

<bibl id="B10">
  <title><p>{Tycho Brahe Parsed Corpus of Historical Portuguese}</p></title>
  <aug>
    <au><snm>Galves</snm><fnm>C</fnm></au>
    <au><snm>Faria</snm><fnm>P</fnm></au>
  </aug>
  <source>URL: http://www.tycho.iel.unicamp.br/corpus/</source>
  <pubdate>2010</pubdate>
</bibl>

<bibl id="B11">
  <title><p>{Merging Comparable Data Sources for the Discrimination of Similar
  Languages: The DSL Corpus Collection}</p></title>
  <aug>
    <au><snm>Tan</snm><fnm>L</fnm></au>
    <au><snm>Zampieri</snm><fnm>M</fnm></au>
    <au><snm>Ljube{\v{s}}ic</snm><fnm>N</fnm></au>
    <au><snm>Tiedemann</snm><fnm>J</fnm></au>
  </aug>
  <source>Proceedings of the 7th Workshop on Building and Using Comparable
  Corpora (BUCC)</source>
  <pubdate>2014</pubdate>
</bibl>

<bibl id="B12">
  <title><p>{Probabilistic Part-of-Speech Tagging Using Decision
  Trees}</p></title>
  <aug>
    <au><snm>Schmid</snm><fnm>H.</fnm></au>
  </aug>
  <source>Proceedings of International Conference on New Methods in Language
  Processing</source>
  <publisher>Manchester, UK</publisher>
  <pubdate>1994</pubdate>
</bibl>

<bibl id="B13">
  <title><p>{Twenty-first Century Corpus Workbench: Updating a Query
  Architecture for the New Millennium}</p></title>
  <aug>
    <au><snm>Evert</snm><fnm>S</fnm></au>
    <au><snm>Hardie</snm><fnm>A</fnm></au>
  </aug>
  <source>Proceedings of Corpus Linguistics</source>
  <pubdate>2011</pubdate>
</bibl>

<bibl id="B14">
  <title><p>{The IMS Corpus Workbench: Corpus Query Processor CQP: User's
  Manual}</p></title>
  <aug>
    <au><snm>Christ</snm><fnm>O</fnm></au>
    <au><snm>Schulze</snm><fnm>BM</fnm></au>
    <au><snm>Hofmann</snm><fnm>A</fnm></au>
    <au><snm>Koenig</snm><fnm>E</fnm></au>
  </aug>
  <source>University of Stuttgart</source>
  <pubdate>1999</pubdate>
</bibl>

<bibl id="B15">
  <title><p>CQPweb—combining power, flexibility and usability in a corpus
  analysis tool</p></title>
  <aug>
    <au><snm>Hardie</snm><fnm>A</fnm></au>
  </aug>
  <source>International journal of corpus linguistics</source>
  <publisher>John Benjamins Publishing Company</publisher>
  <pubdate>2012</pubdate>
  <volume>17</volume>
  <issue>3</issue>
  <fpage>380</fpage>
  <lpage>-409</lpage>
</bibl>

<bibl id="B16">
  <title><p>{Natural Language Processing with Python: Analyzing Text with the
  Natural Language Toolkit}</p></title>
  <aug>
    <au><snm>Bird</snm><fnm>S</fnm></au>
    <au><snm>Klein</snm><fnm>E</fnm></au>
    <au><snm>Loper</snm><fnm>E</fnm></au>
  </aug>
  <publisher>Sebastobol (CA): O'Reilly</publisher>
  <pubdate>2009</pubdate>
</bibl>

<bibl id="B17">
  <title><p>{FreeLing e TreeTagger: Um Estudo Comparativo no {\^A}mbito do
  Portugu{\^e}s}</p></title>
  <aug>
    <au><snm>Gamallo</snm><fnm>P</fnm></au>
    <au><snm>Garcia</snm><fnm>M</fnm></au>
  </aug>
  <pubdate>2013</pubdate>
</bibl>

<bibl id="B18">
  <title><p>{The CQP Query Language Tutorial}</p></title>
  <aug>
    <au><snm>Evert</snm><fnm>S</fnm></au>
  </aug>
  <source>IMS, University of Stuttgart</source>
  <pubdate>2005</pubdate>
</bibl>

<bibl id="B19">
  <title><p>{Automatic Identification of Language Varieties: The Case of
  Portuguese}</p></title>
  <aug>
    <au><snm>Zampieri</snm><fnm>M</fnm></au>
    <au><snm>Gebre</snm><fnm>BG</fnm></au>
  </aug>
  <source>Proceedings of KONVENS</source>
  <pubdate>2012</pubdate>
  <fpage>233</fpage>
  <lpage>-237</lpage>
</bibl>

<bibl id="B20">
  <title><p>Classifying Pluricentric Languages: Extending the Monolingual
  Model</p></title>
  <aug>
    <au><snm>Zampieri</snm><fnm>M</fnm></au>
    <au><snm>Gebre</snm><fnm>BG</fnm></au>
    <au><snm>Diwersy</snm><fnm>S</fnm></au>
  </aug>
  <source>Proceedings of The Fourth Swedish Language Technology Conference
  (SLTC)</source>
  <pubdate>2012</pubdate>
  <fpage>79</fpage>
  <lpage>-80</lpage>
</bibl>

<bibl id="B21">
  <title><p>{Using Bag-of-words to Distinguish Similar languages: How Efficient
  are They?}</p></title>
  <aug>
    <au><snm>Zampieri</snm><fnm>M</fnm></au>
  </aug>
  <source>Proceedings of the 14th International Symposium on Computational
  Intelligence and Informatics (CINTI)</source>
  <pubdate>2013</pubdate>
  <fpage>37</fpage>
  <lpage>-41</lpage>
</bibl>

<bibl id="B22">
  <title><p>{VarClass: An Open-source Language Identification Tool for Language
  Varieties}</p></title>
  <aug>
    <au><snm>Zampieri</snm><fnm>M</fnm></au>
    <au><snm>Gebre</snm><fnm>BG</fnm></au>
  </aug>
  <source>Proceedings of Language Resources and Evaluation (LREC)</source>
  <pubdate>2014</pubdate>
  <fpage>3305</fpage>
  <lpage>-3308</lpage>
</bibl>

<bibl id="B23">
  <title><p>{A Admira{\c{c}}{\~a}o {\`a} Luz dos Corpos}</p></title>
  <aug>
    <au><snm>Santos</snm><fnm>D</fnm></au>
    <au><snm>Mota</snm><fnm>C</fnm></au>
  </aug>
  <source>Oslo Studies in Language</source>
  <pubdate>2015</pubdate>
  <volume>7</volume>
  <issue>1</issue>
</bibl>

<bibl id="B24">
  <title><p>{A Contribution to the Study of the language of Emotion in English
  and Portuguese}</p></title>
  <aug>
    <au><snm>Maia</snm><fnm>BMHS</fnm></au>
  </aug>
  <source>PhD thesis</source>
  <publisher>University of Porto</publisher>
  <pubdate>1994</pubdate>
</bibl>

<bibl id="B25">
  <title><p>{Building a Corpus-based Historical Portuguese Dictionary:
  Challenges and Opportunities}</p></title>
  <aug>
    <au><snm>Junior</snm><fnm>AC</fnm></au>
    <au><snm>Alu{\'\i}sio</snm><fnm>SM</fnm></au>
  </aug>
  <source>TAL</source>
  <pubdate>2009</pubdate>
  <volume>50</volume>
  <issue>2</issue>
  <fpage>73</fpage>
  <lpage>-102</lpage>
</bibl>

<bibl id="B26">
  <title><p>{Procorph: um Sistema de Apoio {\`a} Cria{\c{c}}{\~a}o de
  Dicion{\'a}rios Hist{\'o}ricos}</p></title>
  <aug>
    <au><snm>Candido Jr</snm><fnm>A</fnm></au>
    <au><snm>Alu{\'\i}sio</snm><fnm>SM</fnm></au>
  </aug>
  <source>Proceedings of the VI Workshop Tecnologias da Informa{\c{c}}{\~a}o e
  da Linguagem Humana (TIL)</source>
  <pubdate>2008</pubdate>
  <fpage>1</fpage>
  <lpage>-6</lpage>
</bibl>

<bibl id="B27">
  <title><p>{The Parsing System ``Palavras'': Automatic Grammatical Analysis of
  Portuguese in a Constraint Grammar Framework}</p></title>
  <aug>
    <au><snm>Bick</snm><fnm>E</fnm></au>
  </aug>
  <publisher>Arhus, Denmark: Aarhus Universitetsforlag</publisher>
  <pubdate>2000</pubdate>
</bibl>

<bibl id="B28">
  <title><p>{Grammatical Annotation of Historical Portuguese: Generating a
  Corpus-Based Diachronic Dictionary}</p></title>
  <aug>
    <au><snm>Bick</snm><fnm>E</fnm></au>
    <au><snm>Zampieri</snm><fnm>M</fnm></au>
  </aug>
  <source>Proceedings of Text, Speech, and Dialogue (TSD)</source>
  <pubdate>2016</pubdate>
  <fpage>3</fpage>
  <lpage>-11</lpage>
</bibl>

<bibl id="B29">
  <title><p>{Quantifying the Diachronic Productivity of Irregular Verbal
  Patterns in Romance}</p></title>
  <aug>
    <au><snm>Tang</snm><fnm>K</fnm></au>
    <au><snm>Nevins</snm><fnm>A</fnm></au>
  </aug>
  <source>UCL Working Papers in Linguistics</source>
  <publisher>UCL Working Papers in Linguistics</publisher>
  <pubdate>2013</pubdate>
  <volume>25</volume>
  <fpage>289</fpage>
  <lpage>-308</lpage>
</bibl>

<bibl id="B30">
  <title><p>{The Rise and Fall of the L-shaped Morphome: Diachronic and
  Experimental Studies}</p></title>
  <aug>
    <au><snm>Nevins</snm><fnm>A</fnm></au>
    <au><snm>Rodrigues</snm><fnm>C</fnm></au>
    <au><snm>Tang</snm><fnm>K</fnm></au>
  </aug>
  <source>Probus</source>
  <pubdate>2015</pubdate>
  <volume>27</volume>
  <issue>1</issue>
  <fpage>101</fpage>
  <lpage>-155</lpage>
</bibl>

<bibl id="B31">
  <title><p>{Stylistic Changes for Temporal Text Classification}</p></title>
  <aug>
    <au><snm>{\v{S}}tajner</snm><fnm>S</fnm></au>
    <au><snm>Zampieri</snm><fnm>M</fnm></au>
  </aug>
  <source>Proceeding of Text, Speech and Dialogue (TSD)</source>
  <pubdate>2013</pubdate>
  <fpage>519</fpage>
  <lpage>-526</lpage>
</bibl>

<bibl id="B32">
  <title><p>{Temporal Text Ranking and Automatic Dating of Texts}</p></title>
  <aug>
    <au><snm>Niculae</snm><fnm>V</fnm></au>
    <au><snm>Zampieri</snm><fnm>M</fnm></au>
    <au><snm>Dinu</snm><fnm>LP</fnm></au>
    <au><snm>Ciobanu</snm><fnm>AM</fnm></au>
  </aug>
  <source>Proceedings of EACL</source>
  <pubdate>2014</pubdate>
  <fpage>17</fpage>
  <lpage>-21</lpage>
</bibl>

<bibl id="B33">
  <title><p>{Modeling Language Change in Historical Corpora: The Case of
  Portuguese}</p></title>
  <aug>
    <au><snm>Zampieri</snm><fnm>M</fnm></au>
    <au><snm>Malmasi</snm><fnm>S</fnm></au>
    <au><snm>Dras</snm><fnm>M</fnm></au>
  </aug>
  <source>Proceedings of Language Resources and Evaluation (LREC)</source>
  <pubdate>2016</pubdate>
  <fpage>4098</fpage>
  <lpage>-4104</lpage>
</bibl>

<bibl id="B34">
  <title><p>{The Varitext Platform and the Corpus des Vari{\'e}t{\'e}s
  Nationales du Fran{\c{c}}ais (CoVaNa-FR) as Resources for the Study of French
  from a Pluricentric Perspective}</p></title>
  <aug>
    <au><snm>Diwersy</snm><fnm>S</fnm></au>
  </aug>
  <source>Proceedings of the First Workshop on Applying NLP Tools to Similar
  Languages, Varieties and Dialects (VarDial)</source>
  <pubdate>2014</pubdate>
</bibl>

<bibl id="B35">
  <title><p>{A Computational Approach to the Study of Portuguese Newspapers
  Published in Macau}</p></title>
  <aug>
    <au><snm>Zampieri</snm><fnm>M</fnm></au>
    <au><snm>Malmasi</snm><fnm>S</fnm></au>
    <au><snm>Sulea</snm><fnm>OM</fnm></au>
    <au><snm>Dinu</snm><fnm>LP</fnm></au>
  </aug>
  <source>Proceedings of Workshop on Natural Language Processing Meets
  Journalism (NLPMJ)</source>
  <pubdate>2016</pubdate>
  <fpage>47</fpage>
  <lpage>-51</lpage>
</bibl>

<bibl id="B36">
  <title><p>{From Old Texts to Modern Spellings: An Experiment in Automatic
  Normalisation}</p></title>
  <aug>
    <au><snm>Hendrickx</snm><fnm>I</fnm></au>
    <au><snm>Marquilhas</snm><fnm>R</fnm></au>
  </aug>
  <source>JLCL</source>
  <pubdate>2011</pubdate>
  <volume>26</volume>
  <issue>2</issue>
  <fpage>65</fpage>
  <lpage>-76</lpage>
</bibl>

</refgrp>
} 








\end{document}